\documentclass[10pt]{article}
\usepackage{color}
\usepackage{algorithm}
\usepackage{verbatim}
\usepackage{algpseudocode}
\include{a4wide}
\usepackage{graphicx}
\usepackage{amsmath}
\usepackage{amsfonts}
\usepackage{amssymb}
\usepackage{bm}
\usepackage{pdflscape}
\usepackage{url}
\usepackage{a4wide}
\usepackage{natbib}

\newcommand{\w}[1]{\bm{#1}}
\newcommand{\vect}[1]{\bm{#1}}

\newcommand{\BE}[1]{ \left\langle #1 \right\rangle}

\newcommand{\BC}[1]{ \left\{ #1 \right\}}

\newcommand{\sumlow}[1]{\sum\limits_{#1}}

\newcommand{\prodlow}[1]{\prod\limits_{#1}}
\newcommand{\intlim}[3]{\int_{#1}^{#2}{\!d#3}}

\title{Expectation propagation for continuous time \\ stochastic processes}
\usepackage{authblk}
\author[1]{Botond Cseke\thanks{Email: botcse@microsoft.com}}
\author[2,3]{David Schnoerr\thanks{Email: D.B.Schnoerr@sms.ed.ac.uk}}
\author[4]{Manfred Opper\thanks{Email: opperm@cs.tu-berlin.de}}
\author[2]{Guido Sanguinetti\thanks{Email: gsanguin@inf.ed.ac.uk}}

\affil[1]{Microsoft Research, Cambridge, UK}
\affil[2]{School of Informatics, University of Edinburgh, UK}
\affil[3]{School of Biological Sciences, University of Edinburgh, UK}
\affil[4]{Fakult\"at f\"ur Elektrotechnik und Informationstechnik, Technische Universit\"at Berlin, DE}

\begin{document}

\date{}


\maketitle

\begin{abstract}
We consider the inverse problem of reconstructing the posterior measure over the trajectories of a diffusion process from discrete time observations and continuous time constraints. We cast the problem in a Bayesian framework and derive approximations to the posterior distributions of single time marginals using variational approximate inference. 
We then show how the approximation can be extended to a wide class of discrete-state Markov jump processes by making use of the chemical Langevin equation. Our empirical results show that the proposed method is computationally efficient and provides good approximations for these classes of inverse problems.

\end{abstract}


\section{Introduction}\label{SecIntro}

Physical and technological processes frequently exhibit intrinsic stochasticity. The main mathematical framework to describe and reason about such systems is provided by the theory of continuous time (Markovian) stochastic processes. Such processes have been well studied in chemical physics for several decades as models of chemical reactions at very low concentrations \citep[][e.g.]{gardiner1985handbook}. More recently, the theory has found novel and diverse areas of application including systems biology at the single cell level \citep{wilkinson2011stochastic}, ecology \citep{volkov2007patterns} and performance modelling in computer systems \citep{hillston2005compositional}, to name but a few. The popularity of the approach has been greatly enhanced by the availability of efficient and accurate simulation algorithms \citep{gillespie1977exact, gillespie2013}, which permit a numerical solution of medium-sized systems within a reasonable time frame. 

As with most of science, many of the application domains of continuous time stochastic processes are becoming increasingly data-rich, creating a critical demand for inference algorithms which can use data to calibrate the models and analyse the uncertainty in the predictions. This raises new challenges and opportunities for statistics and machine learning, and has motivated the development of several algorithms for efficient inference in these systems. In this paper, we focus on the Bayesian approach, and formulate the inverse problem in terms of obtaining an approximation to a posterior distribution over the stochastic process, given observations of the system and using existing scientific information to build a prior model of the process.

The data scenario which has attracted most attention within the Bayesian framework is the discretely (partially) observed case. In this scenario, the experiment returns noisy observations of the state (or some components) of  the system at a precise set of time points. To proceed with Bayesian inference over the trajectories/parameters of the process, one needs to approximate the likelihood function, i.e. the conditional probability of the observations given the parameters. This is challenging, as likelihood computations are generally analytically intractable for continuous time processes, and has motivated the development of several approximation strategies based on sampling, variational approximations or system approximations \cite[e.g.][]{beskos2006exact,Opper2008, Ruttor2009, VrOpCo15, Golightly2014,zechner2014scalable,georgoulas2016unbiased}. An alternative data scenario which has received much less attention consists of qualitative observations of the trajectory's behaviour. These are not uncommon: for example, in a biochemical experiment, we may observe that a protein level is above a certain detection threshold over a certain interval of time, without being able to precisely quantify its abundance at any time. In a computer science application, observations may be error logs which report whether the system's trajectory has violated e.g. some safety threshold over its run. This type of observations cannot be localised in time, but it is concerned with global properties of the system's trajectories: we term them continuous time constraints.

Evaluating the probability of a system satisfying some specific trajectory constraints is a highly non-trivial problem; in computer science, this is called the {\it model checking} problem (not to be confused with the problem of model checking in statistics, i.e. assessing the statistical fit to a data set). Evaluating this probability as a function of parameters, providing a likelihood for Bayesian inference, is even more challenging. Recent work, based on Gaussian process emulation and optimisation \citep{Bort2013, DImitris}, has provided a practical solution for maximum likelihood parameter identification for small and medium scale systems; however, Gaussian process optimisation can only provide a point estimate of the parameters, and does not provide a posterior measure over the space of trajectories.

In this paper, we present a flexible approximate scheme for posterior inference in a wide class of stochastic processes from both discrete time observations an continuous time observations/trajectory constraints. The method can be seen as an extension to continuous time of the Expectation-Propagation (EP) approximate inference algorithm \citep{OpperWinther2000, Minka2001}. The algorithm was already presented in \cite{Cseke2013} for latent linear diffusion processes; in this paper, we extend that work in several ways. We extend the approach to a wider class of processes, including Markov jump processes (MJP), by applying moment closure of the corresponding chemical Langevin Equation (CLE). Furthermore, we present a novel derivation of the approach based on optimisation of a variational free energy \citep{OpperWinther2005, Heskes2005}. We demonstrate the approach on new numerical examples, demonstrating the effectiveness and efficiency of the approach on realistic models.

\section{Models}

In this paper we consider Bayesian models for diffusion processes that are observed in discrete and continuous time, where the continuous time observation model can be represented in a specific time integral form. This class of models for  continuous time observations can represent a wide range of phenomena such as continuously observed state space models or path constraints.


We consider a diffusion process $\{\w{x}_t\}$ with known dynamics defined on the time interval $[0,1]$. The process $\{\w{x}_{t}\}$ is defined through the stochastic differential equation (SDE) 
\begin{align}\label{OUequation} 
	d\vect{x}_{t} = \w{a}(\w{x}_{t}, t, \w{\theta})dt + \w{b}(\w{x}_{t}, t, \w{\theta})^{1/2}d\vect{W}_{t},
\end{align} 
where $\{\vect{W}_{t}\}$ is the standard Wiener process \citep{gardiner1985handbook} and $\vect{a}(\w{x}_{t}, t, \w{\theta})$ and $\vect{b}(\vect{x}_t, t, \w{\theta})$ are vector and matrix  valued functions respectively with $\vect{b}(t, \w{x}_t, \w{\theta})$ being positive semi-definite for all~$t \in [0,1]$. The functions $\vect{a}(\w{x}_{t}, t,\w{\theta})$ and $\vect{b}(\vect{x}_t, t, \w{\theta})$ are referred to as {\em drift} and  {\em diffusion} functions, respectively. Alternatively, the process $\{\w{x}_{t}\}$ can also be defined through the Markovian transition probabilities satisfying the Fokker-Planck equation 
\begin{align}\label{EqnFP}
	 \partial_t p(\w{x}_{t} \vert \w{x}_{s}) = -\sumlow{i} \partial_{x}[a_{i}(\w{x}_{t}, t, \w{\theta}) p(\w{x}_{t} \vert \w{x}_{s})] 
		+ \frac{1}{2} \sumlow{ij}\partial_{x_i}\partial_{x_j}[b_{ij}(\w{x}_{t}, t, \w{\theta}) p(\w{x}_{t} \vert \w{x}_{s})].
\end{align}
Even though the process does not possess a formulation through density functions (with respect to the Lebesgue measure),  we use the alias $p(\w{x})$ to denote the ``probability density" of the path/trajectory $\{\w{x}_t\}$ in order to be able to symbolically represent and manipulate the process (or its variables) in the Bayesian formalism. Alternatively, we also use this notation as a reference to a process. 

We assume that the  process can be observed (noisily) both at discrete time points and for continuous time intervals; we  use the  $\vect{y}_{i}$ to denote the discrete time observations at times $t_i \in [0,1]$ and $\vect{y}_{t}$ to denote the continuous time observations for all  $t \in [0,1]$.  We  also use $\w{y} = \{ \{\w{y}_i\}, \{ \w{y}_t\}\}$ to shorten notation where necessary. In this paper we consider models where the observation likelihood admits the general formulation 
\begin{align}\label{EqnJoint}
	p(\{\vect{y}_{i}\}, \{\vect{y}_{t}\} \vert \vect{x}, \w{\theta}) \propto \prodlow{i}p(\vect{y}_{i} \vert \vect{x}_{t_i}, \w{\theta}) \times \exp\BC{-\intlim{0}{1}{t} \, U(\vect{y}_t, \vect{x}_{t}, t, \w{\theta})}.
\end{align}
The term $p(\vect{y}_{i} \vert \vect{x}_{t_i},\w{\theta})$  is the conditional probability of the discrete observation $\w{y}_{i}$  given the state of the process at time $t_i$, while the function $U(\vect{y}_t,\vect{x}_{t}, t, \w{\theta})$ is the negative log-likelihood of the continuous time observations (also referred to as  loss function). We choose this class of models because they are sufficiently expressive to model a wide range of phenomena and  because the Markovian structure of the probabilistic model is preserved, thus making Bayesian inference accessible. For example, the well known linear dynamical systems with the continuous time observation model
$
	d\w{y}_t = \w{C}_t \w{x}_t dt + \w{R}_t^{1/2} d\w{W}_t
$
can be be formulated as follows: (1) we choose a linear drift 
$\w{a}(\w{x}_{t}, t, \w{\theta}) = \w{a}_t \w{x}_t$ and a time-only dependent diffusion $\w{b}(\w{x}_{t}, t, \w{\theta}) = \w{b}_t$ (2) we choose the continuous time loss as the quadratic function $U(\w{y}_t, \w{x}_t, t, \theta) = -\w{x}_t^{T}[\w{C}_{t}^{T}\w{R}_{t}^{-1} \w{y}_t^{\prime}] - \w{x}^{T} [\w{C}_{t}^{T}\w{R}_{t}^{-1} \w{C}_t]\w{x}_t/2$. Another class of models where such time integral representations are useful are temporal log Gaussian Cox process models \citep[e.g.][]{Cseke2013, HarelMO15}.


Bayesian inference in diffusion process models is generally understood as the inference of the posterior process 
\begin{align}\label{EqnStatePost}
	p(\vect{x} \vert \{\vect{y}_{{i}}\}, \{\vect{y}_{t}\}, \w{\theta}) \propto p(\vect{x}\vert \w{\theta}) \times
		p(\{\vect{y}_{i}\} \vert \vect{x}, \w{\theta})  \times p(\{ \vect{y}_{t}\} \vert \vect{x}, \w{\theta}),
\end{align}
and model evidence
\begin{align}\label{EqnPostPar}
	p(\w{y} \,\vert \,\w{\theta}) \propto 
		\int \!d \w{x}  \, p(\vect{x} \vert \w{\theta}) \times p(\{\vect{y}_{i}\} \vert \vect{x}, \w{\theta})  \times p(\{ \vect{y}_{t}\} \vert \vect{x}, \w{\theta}).
\end{align}
Clearly, the existence of a closed form specification of the posterior process \eqref{EqnStatePost} as, say, defined by a new pair of drift and diffusion functions, would be desirable. However, this is only possible in a few special cases such as Ornstein-Uhlenbeck/Gaussian-Markov processes with Gaussian discrete time observations and quadratic loss functions. For this reason we aim to approximate the time marginals  $p(\vect{x}_{t} \vert\, \vect{y} , \vect{\theta})$ (referred to as marginals hereafter) and the model evidence \eqref{EqnPostPar}. 

In this paper we address the case where the intractability is mainly due to the likelihood terms; intractability in the prior (e.g. due to nonlinear drift/ diffusion terms) can be handled as long as efficient methods for approximating marginal moments exist (see Section \ref{SecMC}).  We address this problem in two steps. First, we approximate the likelihood terms by terms that have exponential forms such as
\begin{align}\label{EqnExpAppx}
	p(\vect{y}_{i} \vert \vect{x}_{t_i}, \w{\theta}) \approx \exp\{ \w{\xi}_i^{T} \w{f}(\w{x}_{t_i})\}
	\quad
	\text{and}
	\quad
	U(\vect{y}_t, \vect{x}_{t}, t, \w{\theta}) \approx \w{\xi}_t^{T} \w{f}(\w{x}_{t}) + \text{constant}.
\end{align}
Second, we propose approximations to the posterior marginals $p(\vect{x}_{t} \vert\, \vect{y} , \vect{\theta})$ and the model evidence $p(\w{y} \vert \, \w{\theta} )$ given by \eqref{EqnExpAppx}. Here, the choice of the function $\w{f}$ typically follows from the model and the approximation method. For example, when  the prior $p(\w{x} \vert \w{\theta})$ is Gaussian-Markov or when we can obtain good Gaussian approximations to $p(\w{x}_t \vert \w{\theta})$ we choose $\w{f}$ to be linear and quadratic $\w{f}(\w{x}_t) = (\w{x}_t, -\w{x}_t \w{x}_t^{T}/2)$, thus corresponding to a Gaussian.  In some cases, however, the resulting computations can still be intractable and a factorising Gaussian corresponding to $\w{f}(\w{x}_t) = (x_t^j\}, \{-x_t^j x_t^j/2\})$ is chosen to make computations tractable.
Throughout this paper we consider $\w{f}$ to correspond to a multivariate Gaussian, however, to simplify notation and the emphasise that the results hold for any suitably chosen $\w{f}$ or any
restricted class of Gaussian, we opt for this exponential form representation.

\section{Approximate inference}\label{SecVFE}
In this section we introduce an {\em expectation propagation} (EP) based method to approximate the  mar\-gi\-nals~$p(\w{x}_t \vert \w{y}, \w{\theta})$. 
In its original derivation \citep{OpperWinther2000, Minka2001}, expectation propagation is an algorithm 
for approximating an intractable distribution. One assumes that  the distribution 
is written as a product of certain terms, where
typically one term is the prior and the other terms correspond to the likelihoods of  individual observations.
In the approximating distribution, terms are replaced by tractable 'likelihood' proxies. The EP algorithm aims at achieving 
consistency between the moments of the approximating distribution 
and a set of auxiliary distributions. Each auxiliary distribution is obtained by  replacing a {\em single}
likelihood  proxy by its original counterpart. For many problems auxiliary distributions are still tractable.

EP has been applied to dynamical models \citep{Heskes2002, Ypma2005, Barber:2006} in discrete time. However, a generalisation 
to continuous time processes is not straightforward. It is not immediately clear what an 
addition or removing of a 'single' likelihood term at a given time means for the case of 
continuous time observations/constraints. 

We will show in the following that by first applying EP  
to a time discretised process and by taking a subsequent continuous time limit, we obtain a
well defined algorithm, which treats discrete and continuous time likelihoods in different ways. 

Our derivation of EP for stochastic processes will not follow the original derivation of the 
EP algorithm in \citep{Minka2001} but we will use an equivalent free energy approximation instead.
We thereby follow a line of arguments similar to \cite{OpperWinther2005} where the fixed points 
of the EP algorithm were derived as the stationary points of a specific free energy function. 
This strategy is similar to the derivation of belief propagation algorithms from a Bethe free energy
\citep[e.g.][]{yedidia2000generalized, Heskes2003}.

Our approach will proivde us with an approximation to the  log partition function 
\begin{align}\label{EqnLogZ}
    \log Z(\w{\theta}) &= \log p(\w{y} \vert \w{\theta})
    \nonumber
    \\ 
	 &=\log \int\! \!d\w{x} \, p(\w{x})  \exp\Big\{- \int_{0}^{1}dt U(\w{y}_t, \w{x}_{t},t, \w{\theta})\Big\}   \prod_i p (\w{y}_{i} | \w{x}_{t_i}, \w{\theta}),
\end{align}
which normalises the posterior distribution and also approximations to marginal posterior moments
 \begin{align}
    \BE{\w{f}(\w{x}_t)}_{p(\w{x}_t \vert \w{y})} = \int \!d\w{x}_t \: p(\w{x}_t \vert \w{y}) \w{f}(\w{x}_t),
 \end{align}
where $\w{f}$ typically corresponds to moments up to a certain order. However, our approach is sufficiently general to accommodate  other choices of $\w{f}$ as detailed below. 
Note that the integral in \eqref{EqnLogZ} is over the path $\w{x}$.

 {\bf Notation.} Since we focus on approximating $\log p(\w{y} \vert \w{\theta}) $ and the moments $\BE{\w{f}(\w{x}_t)}_{p(\w{x}_t \vert \w{y} ,\w{\theta})} $ for a fixed $\w{\theta}$, we omit $\w{\theta}$ from our notation in the following.  Moreover, we use the shorthand notation  $U(\w{x}_t, t) =   U(\w{y}_t, \w{x}_{t},t, \w{\theta})$ and use $\w{z}_{1} \cdot \w{z}_2  = \w{z}_{1}^{T}\w{z}_2 $  as an alternative notation for vector inner products. 
 
We refer to the exponential family of distributions defined by the sufficient statistic $\w{f}$ as the family  $\mathcal{F}$ of distributions having the form $r_{\w{\xi}}(\w{z}) = \exp\{ \w{\xi}\!\cdot\!\w{f}(\w{z}) - \log Z_f (\w{\xi})\}$ with $\log Z_f (\w{\xi}) = \log \int\! d\w{z}  \exp\{ \w{\xi}\!\cdot\!\w{f}(\w{z}) \}$. We refer to the parameters $\w{\xi}$ as canonical parameters. The first two moments corresponding to $\w{f}(\w{z})$ can be computed as $\BE{\w{f}(\w{z})}_{r_{\w{\xi}}} = \partial_{\w{\xi}} \log Z_{f}(\w{\xi})$ and 
 $\BE{\w{f}(\w{z}), \w{f}(\w{z})}_{r_{\w{\xi}}} = \partial^2_{\w{\xi}} \log Z_{f}(\w{\xi})$.
 To avoid inconsistent notation, we use $\log Z $ to denote the log partition function in~\eqref{EqnLogZ}.
In the following we assume that for any distribution $s(\w{z})$ for which $\BE{\w{f}(\w{z})}_{s}$ exists, there exists a unique canonical parameter vector $\w{\xi}$ such that 
 $\BE{\w{f}(\w{x})}_{s} = \BE{\w{f}(\w{z})}_{r_{\w{\xi}}}$. This canonical parameter can be defined formally as 
\begin{align}\label{EqnProj}
    \w{\xi}  = \mathop{\mathrm{argmin}}_{\w{\xi}}  \mathrm{KL} [\, s(\w{x})\, \rvert\lvert \,\exp\{ \w{\xi} \cdot \w{f}(\w{x}_t) - \log Z_{f}(\w{\xi})\}\,],
\end{align}
where KL denotes the Kullback-Leibler divergence $\mathrm{KL}[s_1(\w{z}) \rvert\lvert s_2(\w{z})] = \BE{\log(s_1(\w{z})/s_2(\w{z})}_{s_1(\w{z})}$. Throughout this paper we assume that given $\BE{\w{f}(\w{z})}_{s}$, we can compute  $\w{\xi}$ efficiently; we denote this by $\w{\xi} = \text{Project}[s(\w{x}); \w{f}]$ and refer to this by as ``moment-to-canonical parameter transformation".

\subsection{Approximating $\log Z$ and the posterior marginal moments}\label{SecVFE}

In order to approximate $\log Z$ in  \eqref{EqnLogZ}, we first introduce an auxiliary approximating 
process
\begin{align}\label{EqnQ}
	q_{\w{\lambda}}(\w{x}) = \frac{1}{Z_{q}(\w{\lambda})} p(\w{x})  \exp\Big\{ \int_{0}^{1}dt \w{\lambda}_{t}\cdot\w{f}(x_t) + \sum_i \w{\lambda}_{i}\cdot\w{f}(x_{t_i}) \Big\},
\end{align}
where the likelihoods are replaced by simpler likelihood proxies.
We assume that  the partition function $\log Z_q({\w{\lambda}})$ and the marginal moments $q_{\w{\lambda}}(\w{x}_t)$ of the process in \eqref{EqnQ} are computationally tractable or can be approximated with reasonable accuracy.
 We will present a suitable approximation method in Section~\ref{SecMC}.
Here the parameter $\w{\lambda} = \{ \{\w{\lambda}_t\}, \{\w{\lambda}_i\} \}$ is a variational parameter to be optimised
later. Using the process \eqref{EqnQ} and its partition function, we can represent 
\eqref{EqnLogZ} as
\begin{align}\label{EqnLogZnew}
	\log Z = &\log \int\! \!d\w{x} \, p(\w{x})  \exp\Big\{ \int_{0}^{1}dt \w{\lambda}_{t}\cdot\w{f}(\w{x}_t) + \sum_i \w{\lambda}_{i}\cdot\w{f}(\w{x}_{t_i}) \Big\} 
	\nonumber
    \\&
	\times \exp\Big\{- \int_{0}^{1}dt U(\w{x}_{t}, t) -  \int_{0}^{1}dt \w{\lambda}_{t}\cdot\w{f}(\w{x}_t) \Big\}  \times   \prod_i p (\w{y}_{i} | \w{x}_{t_i}) e^{- \w{\lambda}_{i}\cdot\w{f}(\w{x}_{t_i})} .
\end{align}
Rewriting this using \eqref{EqnQ} we obtain
\begin{align}\label{EqnLogZexpand}
	\log Z = & \log Z_{q}(\w{\lambda}) \nonumber
	\\ &
	+ \log \Big\langle  \exp\Big\{- \int_{0}^{1}dt U(\w{x}_{t}, t) -  \int_{0}^{1}dt \w{\lambda}_{t}\cdot\w{f}(x_t) \Big\} \times   \prod_i p (\w{y}_{i} | \w{x}_{t_i}) e^{- \w{\lambda}_{i}\cdot\w{f}(\w{x}_{t_i})} \Big\rangle_{q_{\w{\lambda}}}.
\end{align}
This yields an expression of $\log Z$
as the sum of a tractable log--partition function $\log Z_{q}(\w{\lambda})$ and a correction term accounting for the ``error" resulting from replacing the likelihoods by simpler proxies. A popular approach to simplify the correction would be
to use Jensen's inequality in \eqref{EqnLogZnew} and move the expectation from inside to the 
outside of the logarithm. 
This would give the approximating  bound
\begin{align}\label{EqnLogZVAR}
	\log Z \geq & \log Z_{q}(\w{\lambda}) 
	\\ &
    - \int_{0}^{1}dt \big\langle  U(\w{x}_{t}, t) \big\rangle _{q_{\w{\lambda}}} 
      + \sumlow{i} \BE{ \log p (\w{y}_{i} | \w{x}_{t_i})}_{q_{\w{\lambda}}}
       -  \int_{0}^{1}dt \w{\lambda}_{t}\cdot \big\langle  \w{f}(\w{x}_t) \big\rangle_{q_{\w{\lambda}}}
       - \sumlow{i} \w{\lambda}_{i}\cdot \BE{\w{f}(\w{x}_{t_i})}_{q_{\w{\lambda}}}. \nonumber 
\end{align}
This approximation would be followed by an optimisation of the resulting lower bound 
with respect to the variational  parameters $\w{\lambda}$. 
For recent applications to inference in diffusion processes, see \citep{Arch2007, Ala-Luhtala2014, VrOpCo15, Koeppl2015}.

The EP approximation to the expectation term in \eqref{EqnLogZexpand} proceeds in a different way.
 To this end, we define a set of exponential family marginals $\mathcal{E}(\w{\eta}_t) = \{ q_{\w{\eta}_t}(\w{x}_t);\: t \in [0,1]\}$ with $q_{\w{\eta}_t}(\w{x}_t) = \exp \{ \w{\eta}_{t} \cdot \w{f}(\w{x}_t) - \log Z_{f} (\w{\eta}_t) \}$ that satisfy  the marginal moment matching constraints
\begin{align}\label{EqnMM}
     \BE{\w{f}(x_{t})}_{q_{\w{\lambda}}} = \BE{\w{f}(x_{t})}_{q_{\w{\eta}_t }} \: \text{for all} \: t \in [0,1].
\end{align}
Note that for a Gaussian approximation to the posterior process, these would simply be the Gaussian marginal
densities. Following a similar route  as in \cite{OpperWinther2005} in their derivation of EP, 
our goal will be to approximate the intractable average over the process $q_{\w{\lambda}}$ by
a distribution which {\em factorises in time} and is given  
by the product of  the densities $q_{\w{\eta}_t}(\w{x}_t)$. Hence, we retain the 
information about marginal statistics, but loose the dependencies between different time points. 
This type of approximation---approximating the expectation of a product with the product of expectations given certain moment constraints---is used in a variety of successful approximation methods such as expectation propagation in latent Gaussian models \citep{OpperWinther2000, Minka2001} or belief propagation in pairwise graphical models \citep[e.g.][]{yedidia2000generalized}. The approximation we propose here can be viewed as the extension of this approach to continuous time (infinite dimensional) models.
Of course, it is not trivial to define such a factorising density (corresponding to a delta--correlated
process) directly in continuous time. Hence, it will be 
useful to introduce a discretisation in time into slices of size $\Delta t$ first and then proceed to the limit  
$\Delta t \to 0$. We use the Euler discretisation $\int_{0}^{1}dt U(\w{x}_{t}, t) \backsimeq \sum_{k} \Delta t \: U(\w{x}_{k\Delta t}, k\Delta t) $ and 
$ \int_{0}^{1}dt \w{\lambda}_{t}\cdot\w{f}(x_t) \backsimeq \sum_{k} \Delta t \:\w{\lambda}_{k \Delta t}\cdot\w{f}(x_{k \Delta t}) $ and approximate the expectation of a product, see  \eqref{EqnLogZexpand}, as
{\small
\begin{align*}
	 \Big\langle  \prodlow{k} & \exp\Big\{- \Delta t \: U(\w{x}_{k \Delta t}, k \Delta t) -  \Delta t \: \w{\lambda}_{k \Delta t}\cdot\w{f}(\w{x}_{k \Delta t})\Big\}   
    \times   \prodlow{i} p (\w{y}_{i} | \w{x}_{t_i}) e^{- \w{\lambda}_{i}\cdot\w{f}(\w{x}_{t_i})} \Big\rangle_{q_{\w{\lambda}}} 
    \\
    &
    \approx 
    	\prodlow{k} \Big\langle  \exp\Big\{- \Delta t \: U(\w{x}_{k \Delta t}, k \Delta t) -  \Delta t \: \w{\lambda}_{k \Delta t}\cdot\w{f}(x_{k \Delta t})\Big\}   \Big\rangle_{q_{\w{\eta}_{k\Delta t}}} 
    \times   \prod_i  \Big\langle  p (\w{y}_{i} | \w{x}_{t_i}) e^{- \w{\lambda}_{i}\cdot\w{f}(\w{x}_{t_i})} \Big\rangle_{q_{\w{\eta}_{k\Delta t}}}.
\end{align*}
} 
\noindent
Taking the limit $\Delta t \to 0$ results in the approximation 
\begin{align}\label{EqnFactAppx}
    \log & \Big\langle  \exp\Big\{- \int_{0}^{1}dt U(\w{x}_{t}, t) -  \int_{0}^{1}dt \w{\lambda}_{t}\cdot\w{f}(x_t) \Big\}   
    \times   \prod_i p (\w{y}_{i} | \w{x}_{t_i}) e^{- \w{\lambda}_{i}\cdot\w{f}(\w{x}_{t_i})} \Big\rangle_{q_{\w{\lambda}}} 
    \nonumber
    \\
    &\approx 
    - \int_{0}^{1}dt \big\langle  U(\w{x}_{t}, t) \big\rangle _{q_{\w{\eta}_{t}}}  -  \int_{0}^{1}dt \w{\lambda}_{t}\cdot \big\langle  \w{f}(\w{x}_t) \big\rangle_{q_{\w{\eta}_{t}}}
      +  \sumlow{i}  \log    \Big\langle  p (\w{y}_{i} | \w{x}_{t_i}) e^{- \w{\lambda}_{i}\cdot\w{f}(\w{x}_{t_i})} \Big\rangle_{q_{\w{\eta}_{t_i}}},
\end{align}
where the first two terms follow from $ \log  [\BE{\exp \{ -\Delta t   \: U(\w{x}_{t}, t)\}}_{q_{\w{\eta}_{t}}}] \backsimeq  -\Delta t\BE{U(\w{x}_{t}, t)}_{q_{\w{\eta}_{t}}}$.
A~comparison with \eqref{EqnLogZVAR} shows that the first two terms in \eqref{EqnFactAppx} corresponding to 
continuous time likelihoods would equal their counterparts in the variational bound if the
densities $q_{\w{\eta}_t}(\w{x}_t)$ are the correct marginals of the process $q_{\w{\lambda}}(\w{x})$. This
is the case for a Gaussian posterior approximation. However, the second term is different.

By introducing the variables $\w{\eta}_{i} = \w{\eta}_{t_i}  - \w{\lambda}_i $,  $\w{\eta} = \{ \{\w{\eta}_t\}, \{ \w{\eta}_i \} \}$ and applying \eqref{EqnFactAppx} to  \eqref{EqnLogZexpand} we obtain the approximation
\begin{align}\label{EqnVVFE}
\ln Z \approx L(\w{\lambda}, \w{\eta}) \equiv  \: &  \log Z_{q}(\w{\lambda}) +  
	\sumlow{i}  \log  \int\! dx_{t_i} \, p (\w{y}_{i} | \w{x}_{t_i}) e^{ \w{\eta}_{i}\cdot\w{f}(x_{t_i})} - \sumlow{i} \log Z_{f} (\w{\lambda}_i + \w{\eta}_i)
	\nonumber
	\\
	&
	- \int_{0}^{1}dt \big\langle  U(\w{x}_{t}, t) \big\rangle _{q_{\w{\eta}_{t}}}  -  \int_{0}^{1}dt \w{\lambda}_{t}\cdot \big\langle  \w{f}(\w{x}_t) \big\rangle_{q_{\w{\eta}_{t}}},
\end{align}
where we require the marginal moment matching constraints in \eqref{EqnMM} to hold.  
This approximation contains the two sets of parameters $\w{\eta}$ and $\w{\lambda}$. 
Following similar arguments as given in \cite{OpperWinther2005} to derive EP
for Gaussian latent variable models, we
will now argue that it makes sense to optimise the approximation by computing the 
stationary points of $L(\w{\lambda}, \w{\eta})$ with respect to the variation  of these parameters.
In fact, we can show that variation w.r.t. $\w{\lambda}_t$ leads to the moment matching condition 
\eqref{EqnMM}. Since the exact partition function does not depend on $\w{\lambda}$ it also makes
sense to set the variation w.r.t. $\w{\lambda}$ to zero, thereby making the approximation least sensitive 
w.r.t. to variation of $\w{\lambda}$. 

Using straightforward calculus  one can show that
the differentials of \eqref{EqnVVFE} w.r.t. $\w{\eta}_i$ and $\w{\lambda}_i$ are given by
\begin{align} \label{EqnLdiffD}
	\partial_{\w{\eta}_i} L = \BE{\w{f}(\w{x}_{t_i})}_{\tilde{q}_{\w{\eta}_i}} - \BE{\w{f}(\w{x}_{t_i})}_{q_{\w{\eta}_i + \w{\lambda}_i} }
	\qquad
	\partial_{\w{\lambda}_i} L = \BE{\w{f}(\w{x}_{t_i})}_{q_{\lambda}} - \BE{\w{f}(\w{x}_{t_i})}_{q_{\w{\eta}_i + \w{\lambda}_i} }, 
\end{align}
where we use $\tilde{q}_{\w{\eta}_i}$ to denote the distribution
\begin{align}\label{EqnCavity}
\tilde{q}_{\w{\eta}_i} (\w{x}_t)\propto  p (\w{y}_{i} | \w{x}_{t_i}) e^{ \w{\eta}_{i}\cdot\w{f}(x_{t_i})}.
\end{align}
The variations of \eqref{EqnVVFE} w.r.t. $\w{\eta}_t$ and $\w{\lambda}_{t}$ are
\begin{align} \label{EqnLdiffC}
	\delta_{\w{\eta}_t} L = - \partial_{\eta_t} \big\langle  U(\w{x}_{t}, t) \big\rangle _{q_{\w{\eta}_{t}}}   - \partial^2_{\w{\eta}_t} \log Z_{f} (\w{\eta}_{t}) \w{\lambda}_{t}
	\quad
	\text{and}
	\quad
	\delta_{\w{\lambda}_t} L = \BE{\w{f}(\w{x}_{t})}_{q_{\w{\lambda}}} - \BE{\w{f}(\w{x}_{t})}_{q_{\w{\eta}_t}},
\end{align}
where we  make us of the $\BE{\w{f}(\w{x}_{t})}_{q_{\w{\eta}_t}} = \partial_{\w{\eta}_t} \log Z_{f} (\w{\eta}_{t}) $ property of the exponential family distributions.
Note that from $\delta_{\w{\lambda}_t} L  = 0$ we recover the marginal moment matching constraints postulated in \eqref{EqnMM} and $\w{\eta}_{i} = \w{\eta}_{t_i}  - \w{\lambda}_i$ is guaranteed to hold when setting $ \partial_{\w{\lambda}_i} L = 0$ and $\delta_{\w{\lambda}_t} L =0$.

\subsubsection*{Optimisation} 

Except $\delta_{\w{\eta}_t} L = 0$, all other stationary conditions corresponding to  \eqref{EqnLdiffD} and  \eqref{EqnLdiffC} can be viewed as moment matching conditions.  Since $q_{\w{\eta}_t}$ is from the exponential family,
 $\partial_{\w{\eta}_i} L = 0$ and $\delta_{\w{\lambda}_i} L =0$ can be expressed in terms of canonical parameters as 
 \begin{align}\label{EqnMMCC}
   \w{\lambda}_{i}  + \w{\eta}_i= \text{Project}[\tilde{q}_{\w{\eta}_i}(\w{x}_{t_i}); \w{f}]. 
 \quad
 \text{and}
 \quad
 \w{\lambda}_{i}  + \w{\eta}_i= \text{Project}[q_{\w{\lambda}}(\w{x}_{t_i}); \w{f}].
  \end{align}
 We then use \eqref{EqnMMCC}  to define the fixed point updates
\begin{align}\label{EqnFPD}
	\w{\lambda}_{i}^{new} = \text{Project}[\tilde{q}_{\w{\eta}_i}(\w{x}_{t_i}); \w{f}] - \w{\eta}_i
	\quad
	\text{and}
	\quad
	\w{\eta}_{i}^{new} = \text{Project}[q_{\w{\lambda}}(\w{x}_{t_i}); \w{f}] - \w{\lambda}_i.
	\quad	
\end{align}
Similarly, from \eqref{EqnLdiffC}, we obtain updates
\begin{align}\label{EqnFPC}
	\w{\lambda}_{t}^{new} =  -  [ \partial^2_{\w{\eta}_t} \log Z_{f} (\w{\eta}_{t})]^{-1} \partial_{\eta_t} \big\langle  U(\w{x}_{t}, t) \big\rangle _{q_{\w{\w{\eta}}_{t}}} 
	\quad
	\text{and}
	\quad
	\w{\eta}_{t}^{new} = \text{Project}[q_{\w{\lambda}}(\w{x}_{t_i}); \w{f}].
	\quad	
\end{align}

Readers familiar with  the expectation propagation frameworks proposed in \citep{OpperWinther2000}, \citep{Minka2001} and \citep{Heskes2005} can identify  $\w{\lambda}_i$ as the canonical parameters of the term approximations. The  distributions $\tilde{q}_{\w{\eta}_i}$ are the tilted distributions and $\w{\eta}_i$ are  the parameters of the so-called cavity distributions.
The updates in \eqref{EqnFPD} for the discrete time likelihood proxies correspond to expectation propagation updates. The updates in \eqref{EqnFPC} correspond to non-conjugate variational updates \citep{Knowles2011}. A similar fixed point iteration  for latent Gaussian-Markov models has been derived in \cite{Cseke2013} by applying expectation propagation to the Euler discretisation of the posterior process.


\subsection{Moment approximations by moment closures}\label{SecMC}

In order to run the fixed point iteration we need to (approximately) compute the canonical parameters corresponding to the updates in \eqref{EqnFPD} and \eqref{EqnFPC}.  Since we use exponential family  distributions, this simplifies to (approximately) computing  $\BE{\w{f}(\w{x}_t)}_{\tilde{q}_{\w{\eta}_i}}$,  $\BE{U(\w{x}_t, t)}_{q_{\w{\eta}_t}}$, and $\BE{\w{f}(\w{x}_t)}_{q_{\w{\lambda}}}$ and computing the corresponding canonical parameters.

We further assume that good numerical approximation for $\BE{U(\w{x}_t, t)}_{q_{\w{\eta}_t}}$ and  $\BE{\w{f}(\w{x}_t)}_{\tilde{q}_{\w{\eta}_i}}$  exist and the computational bottleneck of the proposed method is the computation of $\BE{\w{f}(\w{x}_{t})}_{q_{\w{\lambda}}}$. If the assumptions for $\BE{U(\w{x}_t, t)}_{q_{\w{\eta}_t}}$ and  $\BE{\w{f}(\w{x}_t)}_{\tilde{q}_{\w{\eta}_i}}$ do not hold (e.g. $\tilde{q}_{\w{\eta}_i}$ is a complicated multivariate density because of $q_{\w{\eta}_t}$), we can relax the problem by choosing a restricted family of approximations $\mathcal{E}(\w{\eta})$ corresponding to ``weaker" sufficient statistics, say,  $\w{f}(\w{x}_t) = (\{x_t^i\}, \{-[x_t^{i}]^2/2\})$.

Now we introduce approximations to $\BE{\w{f}(\w{x}_{t})}_{q_{\w{\lambda}}}$. Due to the Markovian nature of the process $q_{\w{\lambda}}(\w{x})$, the exact marginals $q_{\w{\lambda}}(\w{x}_{t})$ can be expressed in terms of the distributions
$
q_{\w{\lambda}}(\w{x}_{t} \vert \, \w{\lambda}_{i: t_i \leq t}, \w{\lambda}_{s\leq t})
$ 
and the conditional likelihoods 
$
q_{\w{\lambda}}(\w{\lambda}_{i: t_i > t}, \w{\lambda}_{s> t} \,\vert \, \w{x}_{t} )
$ 
of   $\{\w{\lambda}_{i: t_i > t}, \w{\lambda}_{s> t} \}$
as 
\begin{align} \label{EqnDefSmooth}
q_{\w{\lambda}}(\w{x}_t) \propto  q_{\w{\lambda}}(\w{\lambda}_{i: t_i > t}, \w{\lambda}_{s> t} \, \vert \, \w{x}_{t} )\:q_{\w{\lambda}}(\w{x}_{t} \vert \, \w{\lambda}_{i: t_i \leq t}, \w{\lambda}_{s\leq t}),
\end{align}
where we define
\begin{align}\label{EqnQfw}
	q_{\w{\lambda}}(\w{x}_{t} \,\vert \, \w{\lambda}_{i: t_i \leq t}, \w{\lambda}_{s\leq t}) \propto
	\int \!d \w{x}_{s<t} \: p(\w{x}_{s\leq t})  
	\exp\Big\{ \int_{0}^{t}ds \w{\lambda}_{s}\:\cdot\w{f}(x_s) + \sum_{i: t_i \leq t} \w{\lambda}_{i}\cdot\w{f}(x_{t_i}) \Big\}, 
\end{align}
and
\begin{align}\label{EqnQbw}
	q_{\w{\lambda}}(\w{\lambda}_{i: t_i > t}, \w{\lambda}_{s> t} \,\vert \, \w{x}_{t} )
	\propto
	\int \!d \w{x}_{s>t} \: p(\w{x}_{s>t} \, \vert \, \w{x}_t)  
	\exp\Big\{ \int_{t}^{1}ds \w{\lambda}_{s}\:\cdot\w{f}(x_s) + \sum_{i: t_i > t} \w{\lambda}_{i}\cdot\w{f}(x_{t_i}) \Big\}.
\end{align}
Generally, there are two ways to compute \eqref{EqnDefSmooth}:
\begin{itemize}
\item[(i)]  We independently compute  $q_{\w{\lambda}}(\w{x}_{t} \,\vert \, \w{\lambda}_{i: t_i \leq t}, \w{\lambda}_{s\leq t})$ and $q_{\w{\lambda}}(\w{\lambda}_{i: t_i > t}, \w{\lambda}_{s> t} \,\vert\, \w{x}_{t} )$ by combining the solutions of the forward and backward Fokker-Planck equations---corresponding to prior \eqref{EqnFP}--- with iterative Bayesian updates corresponding to the likelihood proxies in \eqref{EqnQfw} and \eqref{EqnQbw}. We then multiply these quantities to obtain the marginals in \eqref{EqnDefSmooth}.
\item[(ii)] Instead of computing  $q_{\w{\lambda}}(\w{\lambda}_{i: t_i > t}, \w{\lambda}_{s> t} \,\vert \,\w{x}_{t} )$ we compute $q_{\w{\lambda}}(\w{x}_t) $ directly by making use of the {\em smoothing equation} for $q_{\w{\lambda}}(\w{x}_t)$ \citep{Striebel1965, Leondes1970}. The smoothing  equation depends on $\w{\lambda}$ only through 
$q_{\w{\lambda}}(\w{x}_{t} \, \vert \, \w{\lambda}_{i: t_i \leq t}, \w{\lambda}_{s\leq t})$ which are computed as in (i). 
\end{itemize}
In the following we use the latter approach. We do this because the approximation method we introduce in the next section is not well defined for  conditional likelihoods like $q_{\w{\lambda}}(\w{\lambda}_{i: t_i > t}, \w{\lambda}_{s> t} \, \vert \, \w{x}_{t} )$.


When the prior process $p(\w{x}_t)$ is linear, that is, 
$
	d\w{x} = \w{a}_t \w{x}_t dt + \w{b}_{t}^{1/2}d\w{W}_t 
$,
and $\w{f}$ is linear and quadratic, the computations result in solving the Kalman-Bucy forward and backward equations \citep[e.g.][]{Sarkka2013}. These are a set ODEs for the mean and covariance of the corresponding Gaussian distributions for which efficient numerical methods exist. However, when the prior $p(\w{x}_t)$ is non-linear we have to resort to approximations as the computational cost of solving the forward/backward Fokker-Plank equations is generally excessive. Most approximations proposed in the literature assume a parametric form for $q_{\w{\lambda}}(\w{x}_{t} \, \vert \, \w{\lambda}_{i: t_i \leq t}, \w{\lambda}_{s\leq t})$ and $q_{\w{\lambda}}(\w{\lambda}_{i: t_i > t}, \w{\lambda}_{s> t} \,\vert \, \w{x}_{t} )$ and derive ODEs for the parameters of the corresponding approximations.
For example, there is a variety of different approximation methods using  (multivariate) Gaussian approximations presented in \cite{sarkka2010continuous}
and \cite{Sarkka2013}.

\subsubsection{Forward moment approximations} \label{SecFW}

As mentioned above, to compute $q_{\w{\lambda}}(\w{x}_{t} \,\vert \, \w{\lambda}_{i: t_i \leq t}, \w{\lambda}_{s\leq t})$ we need to solve a non-linear Fokker Planck equation \eqref{EqnFP} for which generally no analytic solutions are known. However, note that in our approach we only require the moments
$\BE{\w{f}(\w{x_t})}_{q_{\w{\lambda}}}$ and we hence aim at approximating these directly. If we multiply \eqref{EqnFP} with $\w{f}(\w{x}_t)$ and take the expectation, we obtain the following ODEs for the moments $\BE{\w{f}(\w{x}_t)}_{p}$ of the solution $p$ of \eqref{EqnFP}:
\begin{align}\label{EqnFWM}
	\partial_t \BE{f_l(\w{x}_t)}_{p} = \sumlow{j} \BE{a_j(\w{x}_t, t) \partial_{x_j}f_l(\w{x}_t)}_{p} + \frac{1}{2}\sumlow{j,k} \BE{b_{jk}(\w{x}_t, t) \partial_{x_j}\partial_{x_k} f_l(\w{x}_t)}_{p}.
\end{align}
The moments $\BE{\w{f}(\w{x_t})}_{q_{\w{\lambda}}}$ fulfil similar ODEs which additionally take the measurements $\exp\{ \w{\lambda}_i \cdot \w{f}(\w{x}_{t_i})\}$ and $\exp \{ \int \!dt  \w{\lambda}_t \cdot \w{f}(\w{x}_{t})\}$ into account and which are given in Appendix~B. 
Unfortunately, the ODEs in \eqref{EqnFWM} are not closed (equations for the moments of order $n$ depend on higher order moments), leading to an infinite hierarchy of ODEs \citep{gardiner1985handbook}. Nonetheless, there are well established moment-closure methods to approximate the solutions of these ODEs. {One popular class of moment-closure approximations breaks} this infinite hierarchy by expressing moments above a certain order as functions of lower order moments \citep{Goodman1953, Whittle1957, McQuarrie1964, Lakatos2015, Schnoerr2015}. One is thus left with a finite system of coupled ODEs for which efficient numerical integration schemes exist. {In this article we use the \emph{normal} or \emph{cumulant-neglect} moment closure which sets all cumulants above a certain order to zero. Setting all cumulant above order two to zero corresponds to choosing a multivariate Gaussian distribution as approximation \citep{Gomez2007,Goutsias2007,schnoerr2014validity}}.
We then combine the {resulting} equations with iterative Bayesian updates  corresponding  to $\exp\{ \w{\lambda}_i \cdot \w{f}(\w{x}_{t_i})\}$ and $\exp \{ \int \!dt  \w{\lambda}_t \cdot \w{f}(\w{x}_{t})\}$ to approximate {the moments of} $q_{\w{\lambda}}(\w{x}_{t} \vert \, \w{\lambda}_{i: t_i \leq t}, \w{\lambda}_{s\leq t})$. 
We denote these moments as $\hat{\w{\mu}}^{\rm fw}_{t}$   and the {corresponding}  canonical parameters as $\hat{\w{\eta}}^{\rm fw}_{t}$. More details are given in {Appendix}~\ref{SecAlg}.

\subsubsection{Smoothed moment approximations}\label{SecBW}

As detailed above, in order to approximate $q_{\w{\lambda}}(\w{x}_t)$ we can either approximate $q_{\w{\lambda}}(\w{\lambda}_{i: t_i > t}, \w{\lambda}_{s> t} \,\vert\, \w{x}_{t} )$ or approximate $q_{\w{\lambda}}(\w{x}_t)$ directly. To compute $q_{\w{\lambda}}(\w{\lambda}_{i: t_i > t}, \w{\lambda}_{s> t} \,\vert\, \w{x}_{t} )$ we would need to solve a non-linear backward Fokker-Planck equation. Since $q_{\w{\lambda}}(\w{\lambda}_{i: t_i > t}, \w{\lambda}_{s> t} \,\vert\, \w{x}_{t} )$ is not a distribution, we cannot approximate it using moment closure.  However, we can use moment closure on the moment smoothing equations proposed in \cite{Striebel1965} and \cite{Leondes1970} to directly approximate the moments of $q_{\w{\lambda}}(\w{x}_t)$ instead. These equations compute the marginals moments $\BE{\w{f}(\w{x_t})}_{q_{\w{\lambda}}}$  by using the (exact) $q_{\w{\lambda}}(\w{x}_{t} \,\vert \, \w{\lambda}_{i: t_i \leq t}, \w{\lambda}_{s\leq t})$. They read as
\begin{align}
\label{EqnMMCS}		
	\partial_t \BE{{f}_{l}(\vect{x}_t)}_{q_{\w{\lambda}}} = & \:
		\sumlow{j}\BE{{a}_{j}(\w{x}_t,t) \partial_{x_j}{f}_{l}(\w{x}_t)}_{q_{\w{\lambda}}} 
		\nonumber
		\\& \:
		- \sumlow{j, k}\BE{ \partial_{x_j}{f}_{l}(\w{x}_t) \partial_{x_k} b_{jk}(\w{x}_t,t)}_{q_{\w{\lambda}}} 
		- \frac{1}{2}\sumlow{j, k} \BE{b_{jk}(\w{x}_t, t) \partial_{x_j}\partial_{x_k}{f}_{l}(\w{x}_t)}_{q_{\w{\lambda}}}
		\nonumber
		\\& \:	
		- \sumlow{j, k} \BE{b_{jk}(\w{x}_t,t) \partial_{x_j}{f}_{l}(\w{x}_t) \partial_{x_k} \log q_{\w{\lambda}}(\w{x}_{t} \vert \, \w{\lambda}_{i: t_i \leq t}, \w{\lambda}_{s\leq t})
		}_{q_{\w{\lambda}}}. 
\end{align}
{Similar to \eqref{EqnFWM}, this corresponds to an infinite cascade of coupled ODEs. We solve these approximately} by substituting $q_{\w{\lambda}}(\w{x}_{t} \,\vert \, \w{\lambda}_{i: t_i \leq t}, \w{\lambda}_{s\leq t}) \approx  q_{\hat{\w{\eta}}^{\rm fw}_{t}}(\w{x}_t)$---as obtained in the previous section---into \eqref{EqnMMCS} and applying {a corresponding moment closure}. We denote the resulting moments by $\hat{\w{\mu}}_t$, the canonical parameters by $\hat{\w{\eta}}_t$, and the approximation~by~$q_{\hat{\w{\eta}}_t}(\w{x}_t) \approx q_{\w{\lambda}}(\w{x}_t) $. 


Overall, we have introduced two levels of {approximations:} {(i)} an approximation of \eqref{EqnLogZ} using independence assumptions (Section~\ref{SecVFE}); {(ii)}  moment closure  to approximate the moments required by the optimisation {problem resulting from (i)} (Section~\ref{SecMC}). The first level of approximations reduces the inference problem to moment computations, while the second level performs these moment computations approximately by conveniently combining moment closures and iterative Bayesian updates within an exponential family of distributions.

To derive an algorithm, one first has to decide what family of exponential distributions (choice of $\w{f}$) is best for the model and data at hand. Given $\w{f}$, the form of the moment-closure equations for \eqref{EqnFWM} and \eqref{EqnMMCS} can be derived. It is important to point out that one can do moment closure for a wider set of moments than the ones given by $\w{f}$, be it for computational or accuracy reasons. For example, one can opt for a linear and quadratic $\w{f}$ but compute moments up to $4^{th}$ order for better accuracy in approximating $\BE{\w{f}(\w{x}_t)}_{q_{\w{\lambda}}}$. In {Appendix}~\ref{SecAlg} we provide a detailed description of the algorithm we used to implement the 
(approximate) fixed point iteration in \eqref{EqnFPD} and \eqref{EqnFPC}.

The computational complexity of the algorithm scales linearly w.r.t. time (solving ODEs) while the scaling w.r.t. the dimensionality of $\w{x}_t$ can vary according to $\w{f}$. For the linear and quadratic $\w{f}$ (Gaussian approximations)  we consider in this paper the computational complexity of solving the ODEs resulting from the approach presented in Section~\ref{SecMC}  scales cubically w.r.t. the dimensionality of $\w{x}_t$ (matrix multiplications). This computational complexity is similar to the complexity of the method presented in \cite{Sarkka2013}.

\subsection{Extension to Markov jump processes}\label{sec_mjp}\label{subsec_diff}

{We next aim at extending the applicability of the EP method developed in the previous Section to Markov jump processes (MJPs)}. MJPs are used in many scientific disciplines ranging from queueing theory to epidemiology and systems biology. They constitute a convenient framework to model stochastic dynamics in discrete valued processes. Typically these are systems in which a  set of species interact via various stochastic rules. 
The latter are implemented as transitions or ``jumps'' between the discrete states of the system. 
The state of a $N$-dimensional MJP is given by the vector $ \w{n}_t =(n_t^1, \ldots, n_t^N)$ where each $n_t^i$ is typically a  non-negative integer number. 
In this paper we consider MJPs that have a finite set of possible transitions $\w{n} \to \w{n} + \w{S}_r, r=1, \ldots ,R$, and whose rates $g_r(\w{n_t})$ depend only on the current state $\w{n}_t$ of the system. Here, $\w{S}_r$ is the $r$th column vector of the stoichiometric matrix $\w{S}$ characterising the transitions. The single-time marginal distribution $p(\w{n}_t | \w{n}_0)$ of the process with initial state $\w{n}_0$ is known to fulfil the  master equation \citep{gillespie1992rigorous}
\begin{align}\label{cme}
  \partial_t p(\w{n}_t | \w{n}_0)
  & = 
    \sum_{r=1}^R g_r (\w{n}_t - \w{S}_r) p(\w{n}_t- \w{S}_r | \w{n}_0)
    - \sum_{r=1}^R g_r (\w{n}_t ) p(\w{n}_t | \w{n}_0).
\end{align}
The state  component $n_t^i$ could for instance denote the molecule number of the $i$th species in a chemical reaction system. In this case the transitions correspond to chemical reactions between species and \eqref{cme} is called the \emph{chemical master equation} \citep{gillespie1992rigorous, gillespie2013}. Other types of systems that can be described by a master equation of the type in \eqref{cme} are for instance prey-predator \citep{Reichenbach2006} or epidemic systems \citep{Rozhnova2009}. For all but the most simple systems, analytic solutions to the master equation  in \eqref{cme} are not available, and one has to rely on either stochastic simulations or analytic approximations. 

\subsubsection*{Diffusion approximation} 

{We discuss next a popular method that approximates an MJP by a non-linear diffusion process. In the chemical reaction context the equation defining  the diffusion process is often called the \emph{chemical Langevin equation} \citep{gillespie2000, schnoerr2014}. The approximating non-linear diffusion process is of the same form as the diffusion processes considered in the previous sections defined in \eqref{OUequation}. 
The inference method proposed in this paper can hence be readily applied to MJPs by combining it with the diffusion approximation presented here.}

 
 The {diffusion equation} approximating  an MJP described by \eqref{cme} is a diffusion process with drift and diffusion given by \citep{gillespie2000}
\begin{align}\label{cle}
  \w{a}(\w{x}_{t}, t, \w{\theta})
  & = 
     \w{S} \cdot \w{g} (\w{x}_t), \\
\label{cle2}
  \w{b}(\w{x}_{t}, t, \w{\theta})
  & =
    \w{S} \cdot \text{diag}(\w{g} (\w{x}_t)) \cdot  \w{S}^T.
\end{align}
Here $\w{x}_t$ is the continuous real-valued pendant to $\w{n}_t$, $\w{g}(\w{x}_t)=(g_1(\w{x}_t), \ldots, g_R(\w{x}_t))$, where $g_i(\w{x}_t)$ is the rate function of the $i$th reaction, and $\text{diag}(\w{g}(\w{x}_t))$ is the diagonal matrix with $\w{g}(\w{x}_t)$ on the diagonal.  In Section~\ref{results} we apply the proposed EP method to a Lotka-Volterra system modeled as a MJP by combining it with the diffusion approximation presented here.

\section{Discussion and related work}\label{results}

In  \cite{Cseke2013} we propose an expectation propagation method for diffusion process models where the prior process is Gaussian--Markov. In this paper, we extend this method to models with non-nonlinear prior processes.  Here we use expectation propagation only to approximate the likelihood terms. We avoid approximating the prior process by using moment-closure approximations on the  process resulting from the prior and the likelihood approximations. When we choose a Gaussian--Markov prior process, the method proposed in this paper is identical to the one proposed in  \cite{Cseke2013}.

 In \cite{Arch2007} the authors present a  variational approach to approximate non-linear processes with time-only dependent diffusion terms by Orstein-Uhlenbeck/Gaussian-Markov processes. To our knowledge the extension of the approach in \citep{Arch2007} to prior processes with state-dependent diffusion terms is not straightforward since a Gaussian-Markov approximation to the posterior process would lead to an ill-defined variational objective. The approach presented in this paper provides a convenient way to avoid this problem. We only obtain approximations of the posterior marginals instead of a process approximation \citep{Arch2007}, however, we can address inference problems where the diffusion terms are state dependent. In recent work, \cite{Koeppl2015} proposed an alternative variational approach based on an approximating process with fixed marginal laws. This extends the Gaussian approximation of \cite{Arch2007} to cater for cases where Gaussian marginals are not appropriate, e.g. in stochastic reaction networks where concentrations are constrained positive. The constraint on the marginals however considerably limits the flexibility of their algorithm, and requires a considerable amount of user input;  furthermore, it is unclear how accurate the approximation is in general.

There have been may application of EP in various discrete time models, early works include \cite{Heskes2002}, \cite{Ypma2005} and \citep{Barber:2006}. In these papers the joint distribution of the variables (Markov chain) is approximated by using a factorising approximation in EP. Note that this is not identical to variational mean-field \cite{Minka2001}. As mentioned in Section~\ref{SecVFE} their formulation is not straightforward to extend to continuous time and the derivation we present here is a possible way to go around the problem.
In \cite{Nodelma2005} the authors develop an EP algorithm for continuous time Bayesian networks. Their algorithm
can be viewed as a generalisation of belief propagation \citep{yedidia2000generalized} where each variable in belief propagation corresponds to the
path of a single variable in the Bayesian network. The problems they address, like computing the marginal distribution of the whole path of a group of variables (marginalisation over paths), are not directly related to the ones we address in this paper. Their work is similar in sprit to \cite{Opper2008} and \cite{VrOpCo15} (see below).
 
Inference in Markov jump processes from discrete time observations is a well studied problem, with several available algorithms employing sampling, variational or system approximations \cite[][e.g]{Opper2008, Ruttor2009, Golightly2014,zechner2014scalable,georgoulas2016unbiased}. The extension of our  proposed method to MJPs  (Section~\ref{subsec_diff}) can be viewed as an alternative way to do inference for  such models, with the additional capability of performing inference from continuous time observations.

\cite{sarkka2010continuous} and \cite{Sarkka2013} propose a continuous time extension of the popular unscented transformation in \citep{JulierUD00} to obtain Gaussian state space approximations in SDE models with time-only dependent diffusion terms and both non-linear/non-Gaussian discrete and continuous time observation. In \citep{Ala-Luhtala2014} the authors compare these approaches to the variational method in \citep{Arch2007} which they then use to  improve on their smoothing estimates.
 
 
In  a recent work \cite{VrOpCo15}  present a  mean-field variational approximation where they approximate the posterior process with a set of independent univariate Gaussian processes (factorised approximation). The considered model has polynomial drift terms and state-independent diffusion terms and the observations are at discrete time-points. Due to a clever parameterisations (piecewise polynomials) of the mean and the variance function of the variational approximation the dimensionality of the state can scale to thousands.

\section{Examples}\label{results}



 %
\begin{figure}[t]
\centerline{\includegraphics[width=1\textwidth]{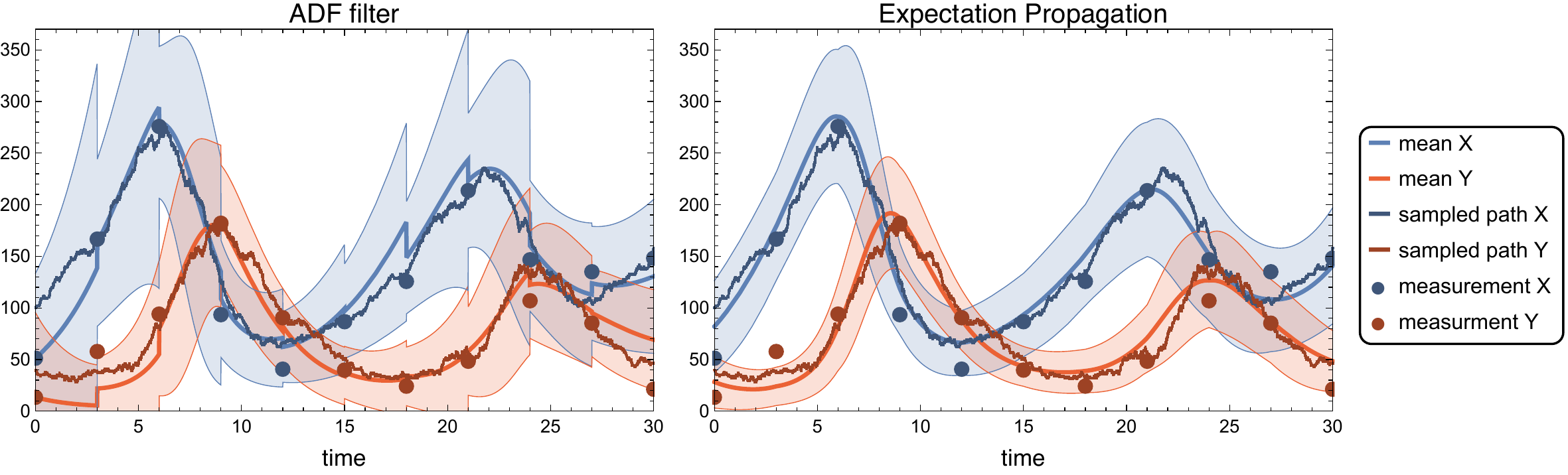}}
\caption{ \small 
EP filtering and smoothing for discrete measurements of the Lotka-Volterra system with reactions in \eqref{lv_reactions}, with log-normal measurement noise with variance $750$.}
\label{fig_lv1}
\end{figure}
%


As an example, we consider a classical benchmark problem, the Lotka-Volterra system. This system is a popular model  describing the nonlinear interactions between a prey and a predator population. The prey $X$ and predator $Y$ interact via the reactions 
\begin{align}\label{lv_reactions}
  \varnothing \xrightarrow{\quad k_0 \quad} X, \quad
  X \xrightarrow{\quad k_1 \quad} X+X,  \quad
  Y+X \xrightarrow{\quad k_2 \quad} Y+Y, \quad Y \xrightarrow{\quad k_3 \quad} \varnothing,
\end{align}
where the first reaction {corresponds} to a birth process of $X$, the second to the reproduction of $X$, the third to reproduction of $Y$ by consumption of one $X$, and the fourth to a death process of $Y$. The corresponding rate function $\w{g}(n_1,n_2)$ and stoichiometric $\w{S}$ matrix can be written as
\begin{align}
  \w{g}(n_1,n_2)
  & =
    (k_0, \ k_1 n_1,\:  k_2 n_1 n_2, \: k_3 n_2)^T, \quad
  \w{S}
  =
    \begin{pmatrix}
      1  &  1 &  -1 & 0 \\
      0  &  0 &  1 & -1   
    \end{pmatrix},
\end{align}
where $n_1$ and $n_2$ are the {number counts} of species $X$ and $Y$, respectively. Depending on the parameters, single realisations of the process show oscillatory behaviour. We choose a fixed parameter set for which this is the case, namely $(k_0,k_1,k_2,k_3) = (5, 0.3, 0.004, 0.6)$. 

{To our knowledge}, no analytic solutions are known for the master equation  \eqref{cme} of this system. To perform approximate inference, we first approximate the master equation  by its diffusion approximation defined in \eqref{cle} and \eqref{cle2}. This allows us to apply the fixed point iteration procedure described in Section~\ref{SecVFE} to perform approximate inference for non-Gaussian likelihoods and continuous time constraints. {The data is generated by simulating the MJP by means of the stochastic simulation algorithm \citep{gillespie1977exact}}. I all scenarios presented in this section, the EP algorithm converged to a accuracy of 0.01 (corresponding to $\tau = 0.01$ in the Appendix~\ref{SecAlg}) after a few iteration, typically 10-20. We have chosen $\tau = 0.01$ because further iterations resulted in no significant changes in the relevant performance measures (see below).

We first consider discrete  time measurements of the system  assuming log-normal measurement noise with a variance of $750$. The panels of  Figure~\ref{fig_lv1} show the sampled data and the inferred approximate state space marginals. The left panel shows the results of the Assumed Density Filtering (ADF) method \citep{Maybeck1982, Minka2001} shown in Algorithm~\ref{AlgADF} of {Appendix}~\ref{SecAlg} while  the right panel shows the results obtained by the expectation propagation (EP) method developed in this paper. The detailed procedure is given in Algorithm~\ref{AlgEP} in {Appendix}~\ref{SecAlg}. 

\begin{figure}[t]
\centerline{\includegraphics[width=1\textwidth]{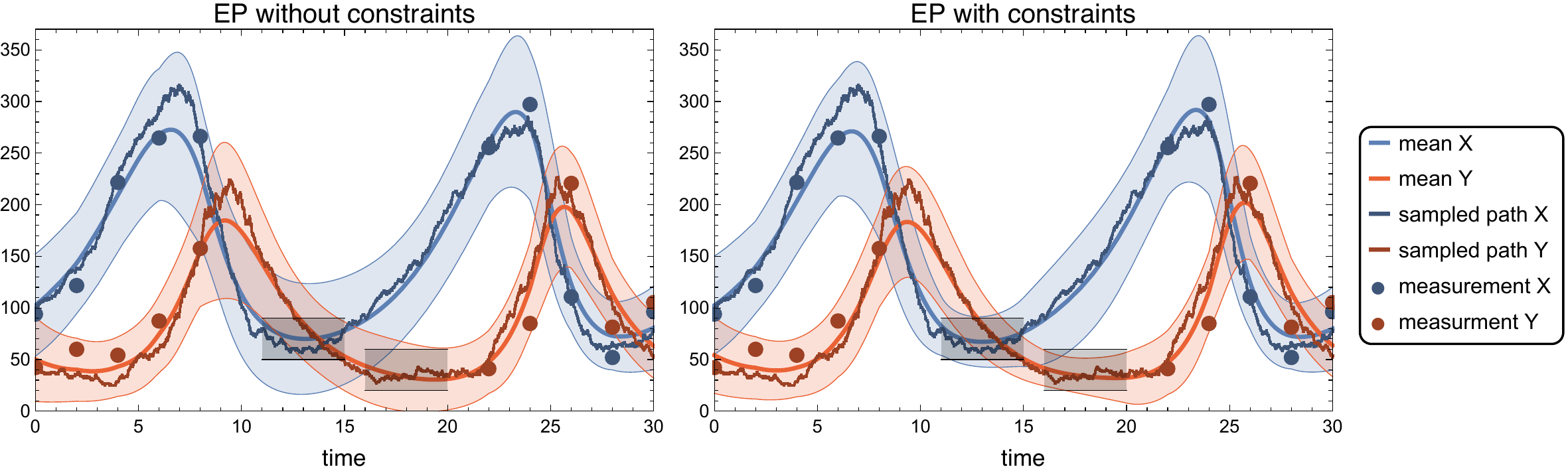}}
\caption{\small  EP for discrete log-normal measurements with variance $750$ and continuous time constraints for the Lotka-Volterra system {defined in} \eqref{lv_reactions}. The panels show the EP result with (right panel) and without (left panel) continuous time constraints taken into account.
}
\label{fig_lv2}
\end{figure}

Next we consider again discrete log-normal measurements with variance $750$  and additionally impose {a continuous time constraint} with loss function
\begin{align}
  U(\w{x}_t), t, \w{\theta})
  & =
    U_1( x^1_t,t, \w{\theta})  + U_2 (x^2_t,  t, \w{\theta}), \quad
    U_{i}(x, t, \w{\theta}) = a_i (x - b_i)^4.
\end{align}
Figure \ref{fig_lv2} shows the results obtained by the EP algorithm, without (left panel) and with (right panel) the  continuous time constraint taken into account. The constraint {was} chosen to limit the process close to its originally sampled path in the regions highlighted on the panel (grey area). {We observe} that the constraints  significantly reduce the variance of the approximate posterior state space marginals in the {corresponding} regions. 

To assess whether  EP improves on ADF, we conducted the following experiment. For several values of the observation noise variance we sampled 40 process paths/trajectories and observation values. We then  measured and averaged the RMSE between the true path and the inferred results using  ADF combined with the corresponding smoothing (ADF-S)---this corresponds to the first step of EP---and the EP algorithm. The latter was iterated until convergence (a few iterations). We computed the RMSE at the data locations (RMSE observations) as well as over the whole sampled path (RMSE path). In this way we assessed both the training and the predictive performance of the approximation. The results are shown in Table~\ref{rms}. We can observe that EP does indeed improve on ADF-S for (almost) all parameter settings. Moreover, it seems that the predictive performance of EP compared to  ADF-S is  slightly increasing  with the increase in the observation noise variance. {This is to be expected, since a larger observation noise variance corresponds to stronger non-Gaussianity of the posterior.}

\begin{table}
\begin{center}
  \begin{tabular}{|c|c|c|c|c|}
  \hline
  \multicolumn{1}{|c} {noise variance} &  \multicolumn{2}{|c} {RMSE observations} &  \multicolumn{2}{|c|} {RMSE path}  \\ \hline
   & ADF & EP  & ADF &  EP \\ \hline \hline
 $ 250 $ & $ 10.23 $ & $ 10.25 $ & $ 11.63 $  & $11.62$  \\ \hline
 $ 500 $ & $ 12.7 $ & $ 12.5 $ & $ 13.5 $  & $13.3$  \\ \hline
 $ 750 $ & $ 15.5 $ & $ 15.0 $ & $16.1 $  & $15.9$ \\ \hline
 $ 1000 $ & $ 16.1 $ & $ 15.9 $ & $ 16.8 $  & $16.5$ \\ \hline
  \end{tabular}
  \caption{\small Comparing RMSE of ADF-S and EP. The table shows the  average  root mean square error (RMSE) resulting {from} the mean of the  approximate {state-space} marginals. The {RMSE} were obtained by averaging over 40 process/data samples for each noise variance value. {We find that the EP method gives slightly more accurate results than ADF.} See {main text} in Section~\ref{results} for further details.}
  \label{rms}
  \end{center}
\end{table}

\section{Conclusions}\label{conclusions}
In this paper, we have derived a novel approximate solution to the Bayesian inference problem for continuous time stochastic processes of diffusion type. This approach can be generalised via a Langevin approximation to Markov jump processes, providing therefore a practical solution to Bayesian inference in a wide class of models. A distinctive feature of our approach is that it can handle both discrete-time observations and trajectory constraints encoded as a continuous-time loss function. The resulting approach is therefore highly flexible. Numerical experiments on a classical benchmark, the Lotka-Volterra system, show both good accuracy and computational performance.

The method we presented is based on a self-consistent algorithm to optimise a variational free energy associated with the inference problem. This formulation is closely related to other variational approaches for inference in continuous-time processes \citep{Arch2007, Cseke2013}, however, it is distinct from others in that we do not seek to reconstruct an approximating process, but focus on computing accurate approximations to the marginal moments of the posterior process. A major advantage of this moment-based approach is that it still leads to a well-defined algorithm even in the case of state-dependent diffusion processes, when a Gaussian variational approach cannot be deployed. 


\bibliographystyle{plainnat}
{\small
\bibliography{ContObsJPA-2206}

\begin{thebibliography}{52}
\providecommand{\natexlab}[1]{#1}
\providecommand{\url}[1]{\texttt{#1}}
\expandafter\ifx\csname urlstyle\endcsname\relax
  \providecommand{\doi}[1]{doi: #1}\else
  \providecommand{\doi}{doi: \begingroup \urlstyle{rm}\Url}\fi

\bibitem[{Ala-Luhtala} et~al.(2014){Ala-Luhtala}, {S{\"a}rkk{\"a}}, and
  {Pich{\'e}}]{Ala-Luhtala2014}
J.~{Ala-Luhtala}, S.~{S{\"a}rkk{\"a}}, and R.~{Pich{\'e}}.
\newblock {Gaussian filtering and variational approximations for Bayesian
  smoothing in continuous-discrete stochastic dynamic systems}.
\newblock \emph{ArXiv e-prints}, 2014.

\bibitem[Archambeau et~al.(2007)Archambeau, Cornford, Opper, and
  Shawe-Taylor]{Arch2007}
C.~Archambeau, D.~Cornford, M.~Opper, and J.~Shawe-Taylor.
\newblock Gaussian process approximations of stochastic differential equations.
\newblock \emph{Journal of Machine Learning Research - Proceedings Track},
  1:\penalty0 1--16, 2007.

\bibitem[Barber(2006)]{Barber:2006}
D.~Barber.
\newblock Expectation correction for smoothed inference in switching linear
  dynamical systems.
\newblock \emph{Journal of Machine Learning Research}, 7:\penalty0 2515--2540,
  December 2006.
\newblock ISSN 1532-4435.

\bibitem[Beskos et~al.(2006)Beskos, Papaspiliopoulos, Roberts, and
  Fearnhead]{beskos2006exact}
A.~Beskos, O.~Papaspiliopoulos, G.~O. Roberts, and P.~Fearnhead.
\newblock Exact and computationally efficient likelihood-based estimation for
  discretely observed diffusion processes (with discussion).
\newblock \emph{Journal of the Royal Statistical Society: Series B (Statistical
  Methodology)}, 68\penalty0 (3):\penalty0 333--382, 2006.

\bibitem[Bortolussi and Sanguinetti(2013)]{Bort2013}
L.~Bortolussi and G.~Sanguinetti.
\newblock Learning and designing stochastic processes from logical constraints.
\newblock In \emph{10th International Conference on Quantitative Evaluation of
  SysTems, QEST 2013}, volume 8054, pages 89--105. Springer Verlag, 2013.

\bibitem[Bortolussi et~al.(2015)Bortolussi, Milios, and Sanguinetti]{DImitris}
L.~Bortolussi, D.~Milios, and G.~Sanguinetti.
\newblock U-check: Model checking and parameter synthesis under uncertainty.
\newblock In \emph{Proceedings of QEST Conference}, 2015.

\bibitem[Csat\'o and Opper(2001)]{CsatoOpper2000}
L.~Csat\'o and M.~Opper.
\newblock Sparse representation for {G}aussian process models.
\newblock In T.~K. Leen, T.~G. Dietterich, and V.~Tresp, editors,
  \emph{Advances in Neural Information Processing Systems 13}, Cambridge, MA,
  USA, 2001. MIT Press.

\bibitem[Cseke et~al.(2013)Cseke, Opper, and Sanguinetti]{Cseke2013}
B.~Cseke, M.~Opper, and G.~Sanguinetti.
\newblock Approximate inference in latent {G}aussian-{M}arkov models from
  continuous time observations.
\newblock In \emph{Advances in Neural Information Processing Systems 26}, pages
  971--979. 2013.

\bibitem[Gardiner(1985)]{gardiner1985handbook}
C.~W. Gardiner.
\newblock \emph{Handbook of stochastic methods}, volume~4.
\newblock Springer Berlin, 1985.

\bibitem[Georgoulas et~al.(2016)Georgoulas, Hilston, and
  Sanguinetti]{georgoulas2016unbiased}
A.~Georgoulas, J.~Hilston, and G.~Sanguinetti.
\newblock Unbiased bayesian inference for population {M}arkov jump processes
  via random truncations.
\newblock \emph{Statistics and Computing}, pages 1--12, 2016.

\bibitem[Gillespie(1977)]{gillespie1977exact}
D.~T. Gillespie.
\newblock Exact stochastic simulation of coupled chemical reactions.
\newblock \emph{The Journal of Physical Chemistry}, 81\penalty0 (25):\penalty0
  2340--2361, 1977.

\bibitem[Gillespie(1992)]{gillespie1992rigorous}
D.~T. Gillespie.
\newblock A rigorous derivation of the chemical master equation.
\newblock \emph{Physica A: Statistical Mechanics and its Applications},
  188\penalty0 (1):\penalty0 404--425, 1992.

\bibitem[Gillespie(2000)]{gillespie2000}
D.~T. Gillespie.
\newblock The chemical {L}angevin equation.
\newblock \emph{The Journal of Chemical Physics}, 113\penalty0 (1):\penalty0
  297--306, 2000.

\bibitem[Gillespie et~al.(2013)Gillespie, Hellander, and
  Petzold]{gillespie2013}
D.~T. Gillespie, A.~Hellander, and L.~R. Petzold.
\newblock Perspective: Stochastic algorithms for chemical kinetics.
\newblock \emph{The Journal of Chemical Physics}, 138\penalty0 (17):\penalty0
  170901, 2013.

\bibitem[Golightly et~al.(2014)Golightly, Henderson, and
  Sherlock]{Golightly2014}
A.~Golightly, D.~A. Henderson, and C.~Sherlock.
\newblock Delayed acceptance particle {MCMC} for exact inference in stochastic
  kinetic models.
\newblock \emph{Statistics and Computing}, pages 1--17, 2014.

\bibitem[Gomez-Uribe and Verghese(2007)]{Gomez2007}
C.~A Gomez-Uribe and G.~C. Verghese.
\newblock Mass fluctuation kinetics: Capturing stochastic effects in systems of
  chemical reactions through coupled mean-variance computations.
\newblock \emph{The Journal of Chemical Physics}, 126\penalty0 (2):\penalty0
  024109, 2007.

\bibitem[Goodman(1953)]{Goodman1953}
L.~A Goodman.
\newblock Population growth of the sexes.
\newblock \emph{Biometrics}, 9\penalty0 (2):\penalty0 212--225, 1953.

\bibitem[Goutsias(2007)]{Goutsias2007}
J.~Goutsias.
\newblock Classical versus stochastic kinetics modeling of biochemical reaction
  systems.
\newblock \emph{Biophysical Journal}, 92\penalty0 (7):\penalty0 2350--2365,
  2007.

\bibitem[Harel et~al.(2015)Harel, Meir, and Opper]{HarelMO15}
Y.~Harel, R.~Meir, and M.~Opper.
\newblock A tractable approximation to optimal point process filtering:
  Application to neural encoding.
\newblock In \emph{Advances in Neural Information Processing Systems 2015},
  pages 1603--1611, 2015.

\bibitem[Heskes(2003)]{Heskes2003}
T.~Heskes.
\newblock Stable fixed points of loopy belief propagation are minima of the
  {B}ethe free energy.
\newblock In \emph{Advances in Neural Information Processing Systems 15}, pages
  359--366, Cambridge, MA, 2003. The MIT Press.

\bibitem[Heskes and Zoeter(2002)]{Heskes2002}
T.~Heskes and O.~Zoeter.
\newblock Expectation propagation for approximate inference in dynamic
  {B}ayesian networks.
\newblock In \emph{Proceedings of the Eighteenth conference on Uncertainty in
  artificial intelligence}, pages 216--223. Morgan Kaufmann Publishers Inc.,
  2002.

\bibitem[Heskes et~al.(2005)Heskes, Opper, Wiegerinck, Winther, and
  Zoeter]{Heskes2005}
T.~Heskes, M.~Opper, W.~Wiegerinck, O.~Winther, and O.~Zoeter.
\newblock Approximate inference techniques with expectation constraints.
\newblock \emph{Journal of Statistical Mechanics: Theory and Experiment},
  2005\penalty0 (11):\penalty0 P11015, 2005.

\bibitem[Hillston(2005)]{hillston2005compositional}
J.~Hillston.
\newblock \emph{A compositional approach to performance modelling}, volume~12.
\newblock Cambridge University Press, 2005.

\bibitem[Julier et~al.(2000)Julier, Uhlmann, and Durrant-Whyte]{JulierUD00}
S.~Julier, J.~K. Uhlmann, and H.~F. Durrant-Whyte.
\newblock A new method for the nonlinear transformation of means and
  covariances in filters and estimators.
\newblock \emph{IEEE Trans. Automat. Contr.}, 45\penalty0 (3):\penalty0
  477--482, 2000.

\bibitem[Knowles and Minka(2011)]{Knowles2011}
D.~A. Knowles and T.~Minka.
\newblock {Non-conjugate Variational Message Passing for Multinomial and Binary
  Regression}.
\newblock In \emph{Advances in Neural Information Processing Systems}, pages
  1701--1709, 2011.

\bibitem[Lakatos et~al.(2015)Lakatos, Ale, Kirk, and Stumpf]{Lakatos2015}
E.~Lakatos, A.~Ale, P.~D.~W. Kirk, and M.~P.~H. Stumpf.
\newblock Multivariate moment closure techniques for stochastic kinetic models.
\newblock \emph{The Journal of Chemical Physics}, 143\penalty0 (9):\penalty0
  094107, 2015.

\bibitem[Lauritzen(1992)]{Lauritzen1992}
S.~L. Lauritzen.
\newblock Propagation of probabilities, means, and variances in mixed graphical
  association models.
\newblock \emph{Journal of the American Statistical Association}, 87\penalty0
  (420):\penalty0 1098--1108, 1992.

\bibitem[Leondes et~al.(1970)Leondes, Peller, and Stear]{Leondes1970}
C.~Leondes, J.~Peller, and E.~Stear.
\newblock {Nonlinear Smoothing Theory}.
\newblock \emph{IEEE Transactions on Systems Science and Cybernetics},
  6\penalty0 (1):\penalty0 63--71, 1970.

\bibitem[Maybeck(1982)]{Maybeck1982}
P.~S Maybeck.
\newblock \emph{Stochastic models, estimation, and control}, volume~3.
\newblock Academic press, 1982.

\bibitem[McQuarrie et~al.(1964)McQuarrie, Jachimowski, and
  Russell]{McQuarrie1964}
D.~A. McQuarrie, C.~J. Jachimowski, and M.~E. Russell.
\newblock Kinetics of small systems. {II}.
\newblock \emph{The Journal of Chemical Physics}, 40\penalty0 (10):\penalty0
  2914--2921, 1964.

\bibitem[Minka(2001)]{Minka2001}
T.~P. Minka.
\newblock \emph{A family of algorithms for approximate {B}ayesian inference}.
\newblock PhD thesis, MIT, 2001.

\bibitem[Nodelman et~al.(2005)Nodelman, Koller, and Shelton]{Nodelma2005}
U.~Nodelman, D.~Koller, and C.~R. Shelton.
\newblock Expectation propagation for continuous time bayesian networks.
\newblock In \emph{Proceedings of the Twenty-first Conference on Uncertainty in
  Artificial Intelligence}, pages 431--440, 2005.

\bibitem[Opper and Sanguinetti(2008)]{Opper2008}
M.~Opper and G.~Sanguinetti.
\newblock Variational inference for {M}arkov jump processes.
\newblock In \emph{Advances in Neural Information Processing Systems}, pages
  1105--1112, 2008.

\bibitem[Opper and Winther(2000)]{OpperWinther2000}
M.~Opper and O.~Winther.
\newblock Gaussian processes for classification: Mean-field algorithms.
\newblock \emph{Neural Computation}, 12\penalty0 (11):\penalty0 2655--2684,
  2000.

\bibitem[Opper and Winther(2005)]{OpperWinther2005}
M.~Opper and O.~Winther.
\newblock Expectation consistent approximate inference.
\newblock \emph{Journal of Machine Learing Research}, 6:\penalty0 2177--2204,
  2005.

\bibitem[Reichenbach et~al.(2006)Reichenbach, Mobilia, and
  Frey]{Reichenbach2006}
T.~Reichenbach, M.~Mobilia, and E.~Frey.
\newblock Coexistence versus extinction in the stochastic cyclic lotka-volterra
  model.
\newblock \emph{Physical Review E}, 74:\penalty0 051907, 2006.

\bibitem[Rozhnova and Nunes(2009)]{Rozhnova2009}
G.~Rozhnova and A.~Nunes.
\newblock Fluctuations and oscillations in a simple epidemic model.
\newblock \emph{Physical Review E}, 79:\penalty0 041922, 2009.

\bibitem[Ruttor and Opper(2009)]{Ruttor2009}
A.~Ruttor and M.~Opper.
\newblock Efficient statistical inference for stochastic reaction processes.
\newblock \emph{Physical Review Letters}, 103\penalty0 (23):\penalty0 230601,
  2009.

\bibitem[S{\"a}rkk{\"a}(2010)]{sarkka2010continuous}
S.~S{\"a}rkk{\"a}.
\newblock Continuous-time and continuous--discrete-time unscented
  {R}auch--{T}ung--{S}triebel smoothers.
\newblock \emph{Signal Processing}, 90\penalty0 (1):\penalty0 225--235, 2010.

\bibitem[S{\"{a}}rkk{\"{a}} and Sarmavuori(2013)]{Sarkka2013}
S.~S{\"{a}}rkk{\"{a}} and J.~Sarmavuori.
\newblock {Gaussian filtering and smoothing for continuous-discrete dynamic
  systems}.
\newblock \emph{Signal Processing}, 93\penalty0 (2):\penalty0 500--510, 2013.

\bibitem[Schnoerr et~al.(2014{\natexlab{a}})Schnoerr, Sanguinetti, and
  Grima]{schnoerr2014}
D.~Schnoerr, G.~Sanguinetti, and R.~Grima.
\newblock The complex chemical {L}angevin equation.
\newblock \emph{The Journal of Chemical Physics}, 141\penalty0 (2):\penalty0
  024103, 2014{\natexlab{a}}.

\bibitem[Schnoerr et~al.(2014{\natexlab{b}})Schnoerr, Sanguinetti, and
  Grima]{schnoerr2014validity}
D.~Schnoerr, G.~Sanguinetti, and R.~Grima.
\newblock Validity conditions for moment closure approximations in stochastic
  chemical kinetics.
\newblock \emph{The Journal of Chemical Physics}, 141\penalty0 (8):\penalty0
  084103, 2014{\natexlab{b}}.

\bibitem[Schnoerr et~al.(2015)Schnoerr, Sanguinetti, and Grima]{Schnoerr2015}
D.~Schnoerr, G.~Sanguinetti, and R.~Grima.
\newblock Comparison of different moment-closure approximations for stochastic
  chemical kinetics.
\newblock \emph{The Journal of Chemical Physics}, 143\penalty0 (18):\penalty0
  185101, 2015.

\bibitem[Striebel(1965)]{Striebel1965}
C.~T. Striebel.
\newblock {Partial differential equations for the conditional distribution of a
  {M}arkov process given noisy observations}.
\newblock \emph{Journal of Mathematical Analysis and Applications},
  11:\penalty0 151--159, 1965.

\bibitem[Sutter et~al.(2015)Sutter, Ganguly, and Koeppl]{Koeppl2015}
T.~Sutter, A.~Ganguly, and H.~Koeppl.
\newblock {A variational approach to path estimation and parameter inference of
  hidden diffusion processes}.
\newblock \emph{arXiv:1508.00506}, 2015.

\bibitem[Volkov et~al.(2007)Volkov, Banavar, Hubbell, and
  Maritan]{volkov2007patterns}
I.~Volkov, J.~R. Banavar, S.~P. Hubbell, and A.~Maritan.
\newblock Patterns of relative species abundance in rainforests and coral
  reefs.
\newblock \emph{Nature}, 450\penalty0 (7166):\penalty0 45--49, 2007.

\bibitem[Vrettas et~al.(2015)Vrettas, Cornford, and Opper]{VrOpCo15}
M.~D. Vrettas, D.~Cornford, and M.~Opper.
\newblock Variational mean-field algorithm for efficient inference in large
  systems of stochastic differential equations.
\newblock \emph{Physical Review E}, 91:\penalty0 012148, 2015.

\bibitem[Whittle(1957)]{Whittle1957}
P.~Whittle.
\newblock On the use of the normal approximation in the treatment of stochastic
  processes.
\newblock \emph{Journal of the Royal Statistical Society. Series B
  (Methodological)}, pages 268--281, 1957.

\bibitem[Wilkinson(2011)]{wilkinson2011stochastic}
D.~J. Wilkinson.
\newblock \emph{Stochastic modelling for systems biology}.
\newblock CRC press, 2011.

\bibitem[Yedidia et~al.(2000)Yedidia, Freeman, and
  Weiss]{yedidia2000generalized}
J.~S. Yedidia, W.~T. Freeman, and Y.~Weiss.
\newblock Generalized belief propagation.
\newblock In \emph{Advances in Neural Information Processing Systems 13},
  volume~13, pages 689--695, 2000.

\bibitem[Ypma and Heskes(2005)]{Ypma2005}
A.~Ypma and T.~Heskes.
\newblock Novel approximations for inference in nonlinear dynamical systems
  using expectation propagation.
\newblock \emph{Neurocomputing}, 69\penalty0 (1-3):\penalty0 85--99, 2005.

\bibitem[Zechner et~al.(2014)Zechner, Unger, Peter, and
  Koeppl]{zechner2014scalable}
C.~Zechner, S.~Unger, M.and~Pelet, M.~Peter, and H.~Koeppl.
\newblock Scalable inference of heterogeneous reaction kinetics from pooled
  single-cell recordings.
\newblock \emph{Nature methods}, 11\penalty0 (2):\penalty0 197--202, 2014.

\end{thebibliography}
}

\appendix

\section{Algorithms and practical considerations}\label{SecAlg}
In Section 5 we compared our EP method with ADF, and we here give a detailed description of both algorithms.
We present two algorithms: (i) the EP algorithm corresponding to the fixed point iteration in Section~\ref{SecVFE} and (ii) an Assumed Density Filtering (ADF) algorithm \citep{Maybeck1982, Lauritzen1992, CsatoOpper2000, Minka2001} and the ADF-S algorithm that performs an extra smoothing step after ADF.  ADF-S a can be viewed as one single step of the~EP.


Before presenting the algorithms, we  provide details of how moment closure and iterative Bayesian updates are combined to approximate the  moments $\BE{\w{f}(\w{x}_t)}_{q_{\w{\lambda}}}$. Let the moment closures for the filtering \eqref{EqnFWM} and smoothing equation \eqref{EqnMMCS} of $\BE{\w{f}(\w{x}_t)}_{q_{\w{\lambda}}}$ result in 
\begin{align}\label{EqnFWSM}
	d\hat{\w{\mu}}_t^{\rm fw} = \mathcal{M}_{\rm fw}(\hat{\w{\mu}}_t^{{\rm fw} }) dt
	\quad
	\text{and}
	\quad
	d\hat{\w{\mu}}_t = \mathcal{M}_{\rm sm}(\hat{\w{\mu}}_t ; \hat{\w{\mu}}_t^{{\rm fw} }) dt.	
\end{align} 
We add the contribution of the Bayesian updates in the forward computation  by 
\begin{align}\
	d\hat{\w{\mu}}_t^{\rm fw} &= \mathcal{M}_{\rm fw}(\hat{\w{\mu}}_t^{{\rm fw} })dt  + \partial^{2}\log Z_{f}(\text{Project}[\hat{\w{\mu}}_t^{\rm fw}; \w{f}] )\w{\lambda}_t dt, 
	\label{EqnAlgFIlter1}
	\\
	d\hat{\w{\mu}}_{t_i +}^{\rm fw} &= \partial \log Z_{f}(\text{Project}[\hat{\w{\mu}}_{t_i}^{\rm fw}; \w{f}] + \w{\lambda}_i), 
	\label{EqnAlgFIlter2}
\end{align} 
where we use $\text{Project}[\hat{\w{\mu}}_t^{\rm fw}; \w{f}]$ to denote the moment-to-canonical parameter transformation. We use $\partial \log Z_{f}$ as a function to denote the canonical-to-moment transformation, that is, $ \hat{\w{\mu}}_t^{\rm fw} =  \partial \log Z_{f}(\text{Project}[\hat{\w{\mu}}_t^{\rm fw}; \w{f}] )$. The form of the second term in the r.h.s. of \eqref{EqnAlgFIlter1} follows from the continuous time Bayesian updates by applying a Taylor expansion to $\partial \log Z_{f}(\text{Project}[\hat{\w{\mu}}_t^{\rm fw}; \w{f}] + dt \w{\lambda}_t)$.
Smoothing is performed by solving the second equation in \eqref{EqnFWSM} backward in time. The measurements do not have to be incorporated explicitly here, they implicitly enter via the solution for~$\hat{\w{\mu}}_t^{\rm fw}$.

Now that the approximation of $\BE{\w{f}(\w{x}_t)}_{q_{\w{\lambda}}}$ is formally fixed, we turn our attention to formulating the algorithm corresponding to the fixed point iteration defined 
in Section~\ref{SecVFE}.
Algorithm~\ref{AlgEP} shows an implementation of this iteration. We initialise $\w{\lambda}_i$ by choosing a good approximation to $p(\w{y}_i \vert \w{x}_{t_i})$ and $U(\w{x}_t, t)$. When this is not possible we simply set   $\w{\lambda}_i = 0$ and $\w{\lambda}_t=0$. In each step of the algorithm we proceed as follows: (i) approximate the moments $\BE{\w{f}(\w{x}_t)}_{q_{\w{\lambda}}}$ by solving  \eqref{EqnAlgFIlter1}-\eqref{EqnAlgFIlter2} and the smoothing equation in \eqref{EqnFWSM}, (ii) compute $\w{\eta}_i$ and the cavity distribution $\tilde{q}_{\w{\eta}_i}$ and (iii) update $\w{\lambda}_i$ and $\w{\lambda}_t$. In many cases 
$\big\langle  U(\w{x}_{t}, t) \big\rangle _{q_{\hat{\w{\eta}}_{t}}}$ can be expressed as a function of of $\hat{\w{\mu}}_{t}$. Therefore, we choose to compute the differentials w.r.t. $\hat{\w{\mu}}_{t}$ instead of $\hat{\w{\eta}}_{t}$. We perform damped updates of $\w{\lambda}_i$ and $\w{\lambda}_t$ and terminate the iteration when the absolute value of the change in the updates  falls below a specified threshold.

Algorithm~\ref{AlgADF} shows an implementation of the ADF algorithm. In ADF a single forward step is performed. Here the iterative Bayesian updates for the likelihood proxies are performed such that 
$\w{\lambda}_t$ and $\w{\lambda}_i$ represent the current estimates computed using $\hat{\w{\mu}}_{t}^{\rm fw}$.  For the continuous time likelihoods this can be viewed as substituting  the update of $\w{\lambda}_t$ in \eqref{EqnFPC} into \eqref{EqnAlgFIlter1}---line 4 of the algorithm. For the discrete time likelihoods the update can be viewed as choosing $\w{\lambda}_i = 0$ when performing the first and only update. 

\begin{algorithm}[t!]
\caption{\small Expectation Propagation (EP)}
\label{AlgEP}
\begin{algorithmic}[1]
{\small
\Function{ExpectationPropagationForDiffusionProcesses}{$\epsilon, \tau,  K_{max}$}
 \State Choose a convenient $\w{\xi}$ such that $\exp\{ \w{\xi} \cdot \w{f}(\w{z})\}$ is reasonably flat
 \State  Initialise  $\w{\lambda}_t = \text{Project}[\exp\{-U(\w{x}_t, t) + \w{\xi} \cdot \w{f}(\w{x}_t)\}; \w{f}]$
  \State Initialise  $\w{\lambda}_i = \text{Project}[p(\w{y}_i \vert \w{x}_{t_i})\exp\{\w{\xi} \cdot \w{f}(\w{z})\}; \w{f}]$
 \For{$k=1,\dots,K_{max}$}
  \State Compute $\hat{\w{\mu}}_t, \: t \in [0, 1]$ and $\hat{\w{\eta}}_{t_i}$  as described in Sections~\ref{SecFW}~and~\ref{SecBW}
  \State Compute $\w{\eta}_i = \hat{\w{\eta}}_{t_i} - \w{\lambda}_i$
   \State Compute $\w{\lambda}_i^{new} =  \text{Project}[\tilde{q}_{\w{\eta}_i} \w{f}] - \w{\eta}_i $ for all $i$
     \State Compute $	\w{\lambda}_{t}^{new} =  - \partial_{\hat{\w{\mu}}_t} \big\langle  U(\w{x}_{t}, t) \big\rangle _{q_{\hat{\w{\eta}}_{t}}} $ for all $t \in [0,1]$ 
     	\State Update $\w{\lambda}_{t}^{new} = (1-\epsilon)\w{\lambda}_{t} + \epsilon \w{\lambda}_{t}^{new} $,   $\w{\lambda}_{i}^{new} = (1-\epsilon)\w{\lambda}_{i} + \epsilon \w{\lambda}_{t}^{new} $
	\If{$\text{max}( \vert \w{\lambda}_{t}^{new}  - \w{\lambda}_{t} \vert, \vert \w{\lambda}_{i}^{new} - \w{\lambda}_{i} \vert) < \tau$}
		\State break
	\EndIf
 \EndFor
 \State Approximate $\log Z$ as described in Sections~\ref{SecVFE} and \ref{SecMC}
\EndFunction
} 
\end{algorithmic}
\end{algorithm}

\begin{algorithm}[t!]
\caption{\small Assumed Density Filtering (ADF/ADF-S)}
\label{AlgADF}
\begin{algorithmic}[1]
{\small
\Function{AssumedDensityFilteringForDiffusionProcesses}{}
	\State Let $d\hat{\w{\mu}}_t^{\rm fw} = \mathcal{M}_{\rm fw}(\hat{\w{\mu}}_t^{\rm fw}) dt$ be the ODE resulting from the moment closure of \eqref{EqnFWM}
	 \For{$i=1,\dots,n$}
	 	\State Solve $d\hat{\w{\mu}}_t^{\rm fw} = \mathcal{M}_{\rm fw}(\hat{\w{\mu}}_t^{\rm fw}) dt - \partial^2\log Z_{f}(\text{Project}[\hat{\w{\mu}}_t^{\rm fw}; \w{f}]) 
			\partial_{\hat{\w{\mu}}_t^{\rm fw}} \big\langle  U(\w{x}_{t}, t) \big\rangle _{q_{\hat{\w{\eta}}_{t}^{\rm fw}}}  dt$ on $(t_{i-1}, t_{i}]$
		\State  Compute ${\w{\eta}}_{i} = \text{Project}[\hat{\w{\mu}}_{t_i}^{\rm fw}; \w{f}]$ and $\tilde{q}_{\w{\eta}_i}(\w{x}_{t_i})$
	  	\State  Compute $\hat{\w{\mu}}_{t_i+}^{\rm fw} = \BE{\w{f}(\w{x}_{t_i})}_{\tilde{q}_{\w{\eta}_i}(\w{x}_{t_i})} $
	  \EndFor
	  \State (for ADF-S compute $\hat{\w{\mu}}_{t}$  as described in Sections~\ref{SecFW}~and~\ref{SecBW})
\EndFunction
}
\end{algorithmic}
\end{algorithm}

\end{document}